\documentclass[a4paper,12pt]{article}

\usepackage[latin1]{inputenc}
\usepackage{amsmath,hhline,fancybox,pifont}

\usepackage{graphicx}

\usepackage[width=16cm, height=25cm]{geometry}
\usepackage[hyperfootnotes=false,breaklinks,linktocpage]{hyperref}
\hypersetup{colorlinks,citecolor=blue,linkcolor=red,urlcolor=blue}

\newcommand{\ly}{\texttt{Layers}}
\title{LAYERS: Yet another Neural Network toolkit}
\author{Roberto Paredes and Jos\'{e}-Miguel Bened\'{i}}
\date{\today}
\begin{document}
\maketitle

\begin{abstract}
\ly is an open source neural network toolkit aim at providing an easy way to implement modern neural networks. The main user target are students and to this end layers provides an easy scriptting language that can be early adopted. The user has to focus only on design details as network totpology and parameter tunning.
\end{abstract}

\setcounter{page}{1}
\section{Introduction}
\label{sec:intro}

\ly\footnote{\url{https://github.com/RParedesPalacios/Layers}} is a neural network toolkit mainly devoted to the academic use. \ly~ aims at providing an easy and fast way for the students to apply and test the theoretical concepts learned.  With \ly~ the student can implement neural netowrks with fully connected and concolutional layers. Moreover \ly~ provides different data manipulation functions.  One of the main requirements in the design of \ly~toolkit was that it should be flexible, easy to use and with an short learning curve.  In order to achieve this we propose developing a \textit{front-end} based on the definition of a specification language of experiments for \ly~ tool, and the construction of its associated compiler.
 
This front-end allows users to define an experiment by means of a program in this specification language.  Then this front-end verifies that the program meets the lexical, syntactic and semantic constraints of language, and later generates a certain intermediate code that is eventually interpreted by the \ly~tool.  In some ways, functions and methods that constitute the \ly~toolkit may be considered the \textit{back-end} that would allow to run the experiment designed by the user.

Other initiatives in order to ease the implementation of NN are for instace \emph{Keras}\footnote{\url{http://keras.io}} and \emph{Lasagne}\footnote{\url{https://lasagne.readthedocs.io/en/latest/}}, among others. But still these lightweight libraries requiere some skills from the students. On the other hand, programms written in layers language are easy to read and focus mainly on network architechture and parametrization, avoiding any other extra information.

\section{\ly~toolkit}
\label{sec:tool}

\begin{figure}
\begin{center}
\includegraphics[width=0.75\textwidth]{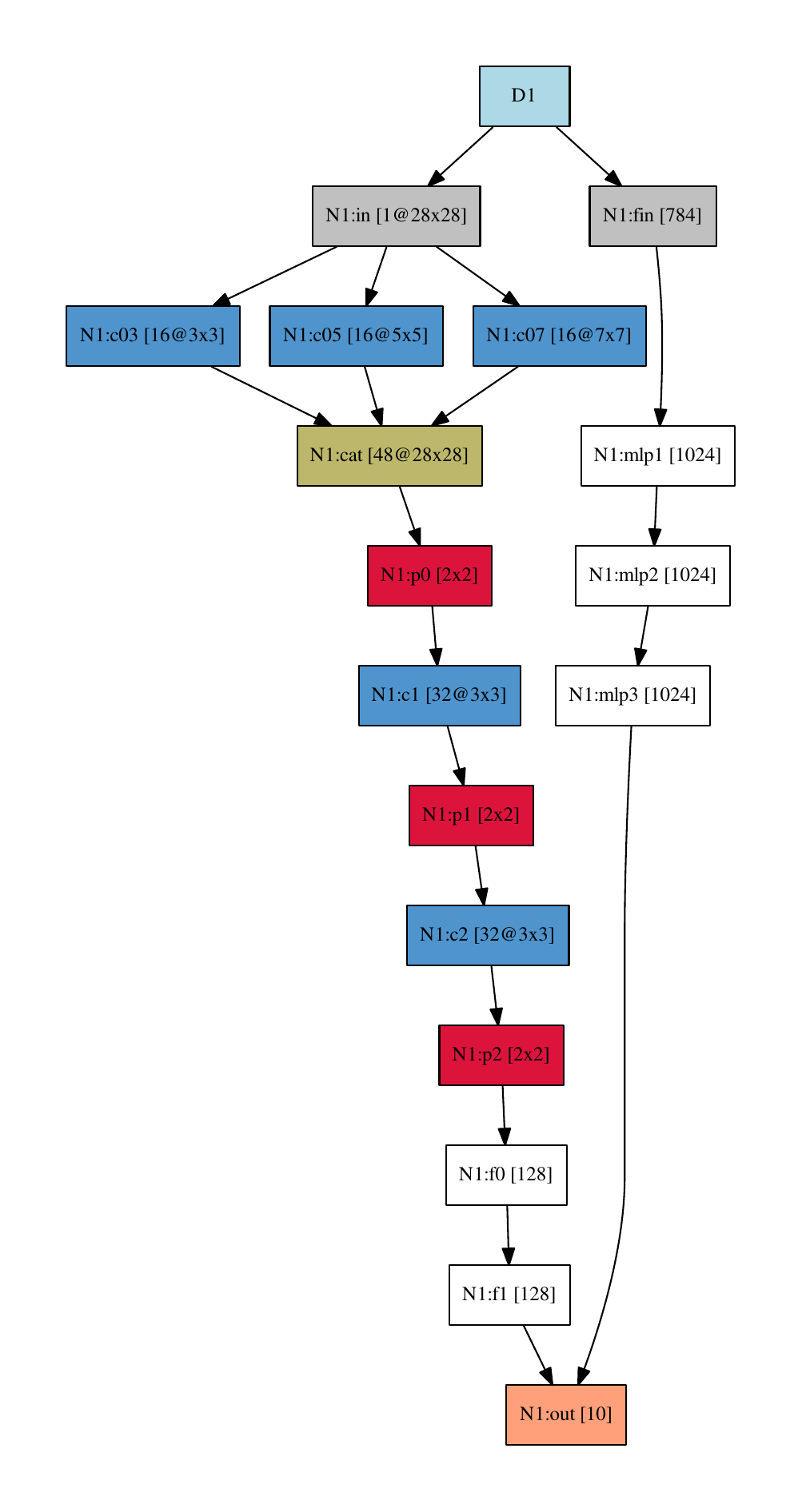}
\end{center}
\caption{An example on Neural Net defined in \ly~}\label{fig:example}
\end{figure}

With \ly~the students can try from very easy neural networks models, e.g. Multi Layer Perceptron, to more complex models with several output layers, multiple connections, semi-supervised learning, combination of convolutional and plain topologies etc. See for instance figure \ref{fig:example}.

And this is the definition of this network:

\begin{figure}
\begin{small}
\begin{verbatim}
network N1 { 
  data tr D1 // Data for training

  // Covolutional Input
  CI in [nz=1, nr=28, nc=28]

  C c03 [nk=16, kr=3, kc=3,rpad=1,cpad=1]	  
  C c05 [nk=16, kr=5, kc=5,rpad=1,cpad=1]	  
  C c07 [nk=16, kr=7, kc=7,rpad=1,cpad=1]	  
  CA cat

  MP p0[sizer=2,sizec=2]
  C c1 [nk=32, kr=3, kc=3]	  	 
  MP p1 [sizer=2,sizec=2]
  C c2 [nk=32, kr=3, kc=3]		
  MP p2 [sizer=2,sizec=2]
  
  // FC reshape
  F   f0 []	
  // FC 
  F  f1 [numnodes=128]
  // Outout
  FO out [classification]

  FI fin // Input fully connected
  F mlp1 [numnodes=1024]
  F mlp2 [numnodes=1024]
  F mlp3 [numnodes=1024]

  // links
  in->c03
  in->c05
  in->c07
  fin->mlp1
  mlp1->mlp2
  mlp2->mlp3
  mlp3->out
  c03->cat
  c05->cat
  c07->cat 
  cat->p0
  p0->c1
  c1->p1
  p1->c2
  c2->p2
  //reshape
  p2->f0
  f0->f1
  f1->out
}
\end{verbatim}
\end{small}
\caption{Example of network definition of figure \ref{fig:example} using \ly~}\label{fig:network}
\end{figure}

\ly~offers two different cost functions, cross-entropy and squared error, for classification and regression problems respectively. Other functions could be added, \ly~is open-source, but with these two cost functions \ly~covers a wide range of academic problems.

\subsection*{\ly~structure}

There are 4 main parts in a \ly~program:
\begin{itemize}
\item Constants
\item Data
\item Networks
\item Scripts
\end{itemize}

Here we describe very basically these blocks, but a much better description can be found in the \ly~tutorial \url{https://github.com/RParedesPalaciotree/master/Tutorial}.
\subsection{Constants}

In the Constants block the user can specify the value of some constants that are used along the \ly~process, batch size, log file and number of threads.

\subsection{Data}

In the Data block the user specify the data objects that can be later linked to networks. These data objects are defined using an associated data file and format (ascii or binary). The data objects have some atrributes and operations that can be acessed in the script block.

\subsection{Networks}

This part is the most important block where the user specify the type of each element (layers) of the neural network and the links among these basic elements, the topology. The elements of the neural networks are mainly two kind of layers: fully connected or convolutional. In this sense we can defined input layers that are fully connected or convolutional, max-pooling layers or cat layers, among others. \ly~have some restrictions in order to define the topology, for instance, a convolutional layer can have more than one \emph{child} layer but only one \emph{parent} layer. But apart from these natural restrictions \ly~provide enough flexibility to define the network topology as can be seen in figure \ref{fig:network}.

\subsection{Scripts}

In the Script block the user can modify the default values of the different objects: data and layers. Morevoer in the script block the user can run fucntions associated to the objects, e.g. can normalize data, run a training for a network, save a network etc.

\section{A front-end for \ly~}
\label{sec:inerfaz}

In the previous Section~\ref{sec:tool} we have presented the most
significant features of the \ly~tool as well as their main alternative
uses.
As it mentioned above, one of the main motivations for the creation of
the \ly~toolkit is that it should be easy to use, and it should have a
learning effort as low as possible.  In order to do this, we propose
here a complete front-end for \ly~toolkit.

This front-end is composed by a specification language of experiments
in \ly~toolkit, and its associated compiler.  

\subsection{\ly~specification language}
\label{subsec:lang}

The \ly~language is a simple specification language for proper
management of this toolkit. A \ly~program defines an experiment or set
of experiments and consists of four main sections: definition of
general constants, data and networks, and description of scripts.

In order to define the \ly~specification language we introduce below the
lexical conventions, and the syntactic-semantic constraints of \ly~.

\subsubsection{Lexical conventions of \ly~}
\label{subsubsec:lex}

Lexical conventions of \ly~language could be summarized as:
\begin{enumerate}
\item The keywords are used to point out the actions, operations and
  general constants. Also they used to define the different parameters
  characterizing the data, the networks or the layers. All keywords
  are reserved, and must be written in lowercase.  Below we show the
  complete list of keywords.

  \vspace{2mm}
  \begin{tabular}{|llllll|}\hline
    $\mathbf{const}$&$\mathbf{batch}$&$\mathbf{threads}$&
    $\mathbf{log}$&$\mathbf{data}$&$\mathbf{filename}$\\

    $\mathbf{ascii}$&$\mathbf{binary}$&$\mathbf{network}$&
    $\mathbf{tr}$&$\mathbf{ts}$&$\mathbf{va}$\\

    $\mathbf{FI}$&$\mathbf{CI}$&$\mathbf{F}$&
    $\mathbf{FO}$&$\mathbf{C}$&$\mathbf{MP}$\\

    $\mathbf{CA}$&$\mathbf{nz}$&$\mathbf{nr}$&
    $\mathbf{nc}$&$\mathbf{cr}$&$\mathbf{cc}$\\

    $\mathbf{numnodes}$&$\mathbf{local}$&$\mathbf{classification}$&
    $\mathbf{regression}$&$\mathbf{autoencoder}$&$\mathbf{nk}$\\

    $\mathbf{kr}$&$\mathbf{kc}$&$\mathbf{rpad}$&
    $\mathbf{cpad}$&$\mathbf{stride}$&$\mathbf{sizer}$\\

    $\mathbf{sizec}$&$\mathbf{script}$&$\mathbf{mu}$&
    $\mathbf{mmu}$&$\mathbf{l2}$&$\mathbf{l1}$\\

    $\mathbf{maxn}$&$\mathbf{drop}$&$\mathbf{noiser}$&
    $\mathbf{noisesd}$&$\mathbf{brightness}$&$\mathbf{contrast}$\\

    $\mathbf{lambda}$&$\mathbf{noiseb}$&$\mathbf{bn}$&
    $\mathbf{act}$&$\mathbf{shift}$&$\mathbf{flip}$\\

    $\mathbf{balance}$&$\mathbf{printkernels}$&$\mathbf{train}$&
    $\mathbf{load}$&$\mathbf{save}$&$\mathbf{testout}$\\

    $\mathbf{zscore}$&$\mathbf{yuv}$&$\mathbf{center}$&
    $\mathbf{div}$& & \\\hline
\end{tabular}

\vspace{4mm}
\item Special symbols are  the following:
\verb+    { }  [  ]  .  ,  =  ->+

\item Identifiers ($\mathbf{id}$), unsigned numerical constants
  ($\mathbf{cte}$), and complete paths to files ($\mathbf{nfile}$) are
  symbols (tokens) whose lexical constraints can be defined by the
  following regular expressions:

  \vspace{-2mm}
  \begin{center}
  \begin{tabular}{l@{ $=$ }l}
  $digit$     & \verb![0-9]!\\[0.5mm] 
  $letter$    & \verb![_a-zA-Z]!\\[0.5mm] 
  $digits$    & \verb!{digit}+!\\[0.5mm] 
  $opfraction$& \verb!(.{digits})?!\\[0.5mm] 
  $\mathbf{id}$  &\verb!{letter}({letter} | {digit})*!
                 \\[0.5mm]
  $\mathbf{cte}$ &\verb!{digits}{opfraction}! \\[0.5mm] 
  $\mathbf{nfile}$&\verb!"([^\0 ])+"!\\[-2mm]
  \end{tabular}
  \end{center}
  Where a file path can be any character enclosed between quotes,
  except the characters null and blank. Lower and uppercase letters
  are distinct.
\item A comment starts with a double slash (\texttt{//}) ~and ends
  with a newline, and they can be placed anywhere white space can
  appear.  Comments may not be nested.
\item White space consists of blanks, newlines, and tabs. White space
  is ignored except that it must separate $\mathbf{id}$'s,
  $\mathbf{cte}$'s, $\mathbf{nfile}$'s, and keywords.
\end{enumerate}
\subsubsection{Syntax and semantics of \ly~}
\label{subsubsec:syn}

A program in \ly~specification language consists of an optional
definition of general constants, followed by a sequence of definitions
of data, networks and scripts, in any order.
\begin{center}
\begin{tabular}{l@{  $\rightarrow$  }l}
$experiment$ & $constants$  ~$definitions$ \\[0.5mm]
$constants$  & $\mathbf{const}$ ~$\{$ ~$lconst$ ~$\}$ ~~$|$ 
               ~~$\epsilon$ \\[0.5mm]
$lconst$     & $lconst$ ~$const$ ~~$|$ ~~$const$ \\[0.5mm]
$const$      & $\mathbf{batch}$ ~$=$ ~$\mathbf{cte}$ ~~$|$  
               ~~$\mathbf{threads}$ ~$=$ ~$\mathbf{cte}$ ~~$|$ 
               ~~$\mathbf{log}$ ~$=$ ~$\mathbf{nfile}$ \\[0.5mm]

$definitions$& $definitions$ ~$def$ ~~$|$ ~~$def$ \\[0.5mm]
$def$        & $data$ ~~$|$ ~$networks$ ~~$|$ ~$scripts$ \\[0.5mm]
\end{tabular}
\end{center}
General constants have the following default values: 
size of the batch for the network ($\mathbf{batch}\,=\,100$), 
number of threads for parallelization ($\mathbf{threads}\,=\,4$), and
log file where some messages are saved 
($\mathbf{log}\,=\,``netparser.log''$).

\medskip %
Data section defines the corpora to be used in experiments.  For its
management in the program, data must be associated with internal
variable names. Data can be read in ascii ($\mathbf{ascii}$) or binary
($\mathbf{binary}$) format, being possible to use their full path.
\begin{center}
\begin{tabular}{l@{  $\rightarrow$  }l}
$data$     & $\mathbf{data}$ ~$\{$ ~$ldata$ ~$\}$\\[0.5mm] 
$ldata$    & $ldata$ ~$datum$  ~~$|$ ~~$datum$\\[0.5mm]
$datum$    & $\mathbf{id}$ ~$[$ ~$ldatumpar$ ~$]$\\[0.5mm]
$ldatumpar$& $ldatumpar$ ~$,$ ~$datumpar$ ~~$|$ ~~$datumpar$\\[0.5mm]
$datumpar$ & $\mathbf{filename}$ ~$=$ ~$\mathbf{nfile}$ 
             ~~$|$ ~~$filetype$ \\[0.5mm]
$filetype$ & $\mathbf{ascii}$  ~~$|$ ~~$\mathbf{binary}$\\[0.5mm]
\end{tabular}
\end{center}

\medskip %
The definition of a network is composed in turn of 3 main parts:
definition of the used data, the selected layers, and connections
defined between these layers.
\begin{center}
\begin{tabular}{lcl}
$networks$   &$\rightarrow$&$\mathbf{network}$ ~$\mathbf{id}$ ~$\{$ 
              ~$netdata$ ~$lstatements$ ~$\}$ \\[0.5mm]
$lstatements$&$\rightarrow$&$lstatements$ ~$statement$ ~~$|$ 
              ~~$statement$\\[0.5mm]
$statement$  &$\rightarrow$&$layer$ ~~$|$ ~~$edge$\\[0.5mm]
$netdata$    &$\rightarrow$&$\mathbf{data}$ ~$\mathbf{tr}$ 
              ~$\mathbf{id}$ ~$rnetdata$\\[0.5mm]
$rnetdata$   &$\rightarrow$&$rnetdata$ ~$fnetdata$ ~~$|$ 
              ~~$\epsilon$\\[0.5mm]
$fnetdata$   &$\rightarrow$&$\mathbf{data}$ ~$\mathbf{va}$ 
              ~$\mathbf{id}$ ~~$|$ ~~$\mathbf{data}$ 
              ~$\mathbf{ts}$ ~$\mathbf{id}$\\[2.5mm]
$layer$      &$\rightarrow$&$\mathbf{FI}$ ~$\mathbf{id}$ \\[0.5mm]
             &$|$&$\mathbf{CI}$ ~$\mathbf{id}$ ~$[$ ~$cilparam$ ~$]$ 
              \\[0.5mm]
             &$|$&$\mathbf{F}$ ~$\mathbf{id}$ ~$[$ ~$fparam$ ~$]$
              \\[0.5mm]
             &$|$&$\mathbf{FO}$ ~$\mathbf{id}$ ~$[$ ~$folparam$ ~$]$
              \\[0.5mm]
             &$|$&$\mathbf{C}$ ~$\mathbf{id}$ ~$[$ ~$clparam$ ~$]$
              \\[0.5mm]
             &$|$&$\mathbf{MP}$ ~$\mathbf{id}$ ~$[$ ~$mplparam$ ~$]$
              \\[0.5mm]
             &$|$&$\mathbf{CA}$ ~$\mathbf{id}$ \\[0.5mm]
$cilparam$   &$\rightarrow$&$cilparam$ ~$,$ ~$ciparam$ ~~$|$ 
              ~~$ciparam$\\[0.5mm]
$ciparam$    &$\rightarrow$&$\mathbf{nz} = \mathbf{cte}$ ~~$|$
                ~~$\mathbf{nr} = \mathbf{cte}$ ~~$|$
                ~~$\mathbf{nc} = \mathbf{cte}$ ~~$|$
                ~~$\mathbf{cr} = \mathbf{cte}$ ~~$|$
                ~~$\mathbf{cc} = \mathbf{cte}$\\[0.5mm]
$fparam$     &$\rightarrow$&$\mathbf{numnodes} = \mathbf{cte}$ ~~$|$
              ~~$\mathbf{local}$ ~~$|$ ~~$\epsilon$ \\[0.5mm]
$folparam$   &$\rightarrow$&$folparam$ ~$foparam$ ~~$|$ 
              ~~~$foparam$ \\[0.5mm]
$foparam$    &$\rightarrow$&$\mathbf{classification}$ ~~$|$ 
              ~~$\mathbf{regression}$ ~~$|$ ~~$\mathbf{autoencoder}$
              \\[0.5mm]
$clparam$    &$\rightarrow$&$clparam$ ~$,$ ~$cparam$ ~~$|$ 
              ~~$cparam$\\[0.5mm]
$cparam$     &$\rightarrow$&$\mathbf{nk} = \mathbf{cte}$ ~~$|$
              ~~$\mathbf{kr} = \mathbf{cte}$ ~~$|$
              ~~$\mathbf{kc} = \mathbf{cte}$  \\[0.5mm]
             &$|$&$\mathbf{rpad} = \mathbf{cte}$ ~~$|$
              ~~$\mathbf{cpad} = \mathbf{cte}$ ~~$|$
              ~~$\mathbf{stride} = \mathbf{cte}$\\[0.5mm]
$mplparam$   &$\rightarrow$&$mplparam$ ~$,$ ~$mpparam$ ~~$|$ 
              ~~$mpparam$\\[0.5mm]
$mpparam$    &$\rightarrow$&$\mathbf{sizer} = \mathbf{cte}$ ~~$|$
                ~~$\mathbf{sizec} = \mathbf{cte}$\\[2.5mm]
$edge$       &$\rightarrow$&$namelayer$ \verb! -> ! ~$namelayer$
              \\[0.5mm]
$namelayer$  &$\rightarrow$& $\mathbf{id}$ ~$.$ ~$\mathbf{id}$ ~~$|$ 
              ~~$\mathbf{id}$\\[0.5mm]
\end{tabular}
\end{center}

\smallskip %
For each network it is necessary to explicitly define the data sets it
uses: training data set ($\mathbf{tr}$), which is mandatory, and
validation ($\mathbf{va}$) and test ($\mathbf{ts}$) data, which are
both optional.  When test or validation data are provided the error
function of the network will be also evaluated for that data sets.
Data identifiers ($\mathbf{id}$) must be previously defined in a data
section.

\smallskip %
As described in the previous Section~\ref{sec:tool}, in the
\ly~toolkit are defined the following types of layers:
\vspace{0mm}
\begin{list}{\textbullet}{%
\setlength{\topsep}{4mm}\setlength{\itemsep}{1mm}}
\item \textit{Input Fully Connected layer} ($\mathbf{FI}$). The FI
  layer has no parameters, just serve as an interface with the input
  data.  The number of units of the layer coincides with the
  dimensionality of the representation of the input data.
\item \textit{Input Covolutional layer} ($\mathbf{CI}$). The CI layer
  has three mandatory parameters that indicate how the raw data have
  to be mapped into an input map: number of channels ($\mathbf{nz}$),
  number of image rows ($\mathbf{nr}$), and number of image cols
  ($\mathbf{nc}$).

  The CI layer also has three optional parameters: ($\mathbf{cr}$)
  crop rows, and ($\mathbf{cc}$) crop cols.  In case $\mathbf{cr}$ and
  $\mathbf{cc}$ parameters are not defined, they take the default
  values of the parameters $\mathbf{nr}$ and $\mathbf{nc}$
  respectively.

\item \textit{Fully Connected layer} ($\mathbf{F}$). The F layer has
  only one mandatory parameter: number of nodes ($\mathbf{numnodes}$).

\item \textit{Ouput layer} ($\mathbf{FO}$). The FO layer has only one
  mandatory parameter indicating the criterion for treating the cost
  error: cross-entropy ($\mathbf{classification}$) , or mean squared
  error ($\mathbf{regression}$).

  Additionally, for the $\mathbf{regression}$ criterion an optional
  parameter of autoencoder ($\mathbf{autoencoder}$) can be defined.

\item \textit{Convolutional layer} ($\mathbf{C}$). The C layer has
  three mandatory parameters that indicate number of kernels
  ($\mathbf{nk}$), height of kernel ($\mathbf{kr}$), and width of kernel
  ($\mathbf{kc}$)

  The C layer also has three optional parameters: $\mathbf{rpad}$ to
  indicate that the method is done in rows, $\mathbf{cpad}$ to
  indicate that the method is done in cols, and $\mathbf{stride}$
  stride value. The default values are respectively $0$, $0$ and $1$.

\item \textit{MaxPooling layer} ($\mathbf{MP}$). The MP layer has two
  mandatory parameter: ($\mathbf{MP}$) height of the pooling region, and
  ($\mathbf{MP}$) width of the pooling region.

\item \textit{Cat layer} ($\mathbf{CA}$). The CA does not require any
  parameter.
\end{list}

The connection between layers is defined by means the operator
(\verb!->!).  Both the source layer as the target layer can be defined
by the simple name of the layer, when there is no ambiguity, or by the
network name followed by a period (\verb!.!) and followed by the name
of the layer.

\medskip %
The definition of scripts is composed of a sequence of actions of two
different types: $amendment$ ~for modifying some parameter of networks,
layers or data, and ~$command$ ~for defining some operations on
networks, layers or or data.
\begin{center}
\begin{tabular}{lcl}
$scripts$  &$\rightarrow$&$\mathbf{script}$ ~$\{$ ~$lactions$ 
            ~$\}$\\[0.5mm] 
$lactions$ &$\rightarrow$&$lactions$ ~$action$ ~~$|$ ~~$action$
            \\[0.5mm] 
$action$   &$\rightarrow$&$amendment$ ~~$|$ ~~$command$\\[1.5mm] 
$amendment$&$\rightarrow$&$\mathbf{id}$ ~$.$ ~$\mathbf{id}$ ~$.$ 
            ~$parameter$ ~~$|$ ~$\mathbf{id}$ ~$.$ ~$parameter$ 
            \\[0.5mm] 
$parameter$&$\rightarrow$&$paramctr$ ~$=$ $\mathbf{cte}$ ~~$|$ 
              ~~$paramcte$ ~$=$ ~$\mathbf{cte}$\\[0.5mm] 
$paramctr$ &$\rightarrow$&$\mathbf{mu}$ ~~$|$ ~~$\mathbf{mmu}$ ~~$|$ 
            ~~$\mathbf{l2}$ ~~$|$ ~~$\mathbf{l1}$ ~~$|$ 
            ~~$\mathbf{maxn}$ ~~$|$ ~~$\mathbf{drop}$ ~~$|$ 
            ~~$\mathbf{noiser}$ \\[0.5mm]
           &$|$& $\mathbf{noisesd}$ ~~$|$~~$\mathbf{noiseb}$ ~~$|$ 
            ~~$\mathbf{brightness}$ ~~$|$ ~~$\mathbf{contrast}$ ~~$|$ 
            ~~$\mathbf{lambda}$ \\[0.5mm]
$paramcte$ &$\rightarrow$&$\mathbf{bn}$ ~~$|$ ~~$\mathbf{act}$ ~~$|$ 
              ~~$\mathbf{shift}$ ~~$|$ ~~$\mathbf{flip}$
              ~~$|$ ~~$\mathbf{balance}$\\[0.5mm]
\end{tabular}
\end{center}
\smallskip %
Whether for a particular layer or for all layers of a network we can
modify some of their parameters.  For layers and networks, the integer
parameters that can currently be modified they are:

\vspace{0mm}
\begin{list}{\textbullet}{%
\setlength{\topsep}{1mm}\setlength{\itemsep}{-1mm}}
\item $\mathbf{bn}$, batch normalization ($\{0, 1\}$)
\item $\mathbf{act}$, activation (0 Linear, 1 Relu, 2 Sigmoid, 3 ELU)
\item $\mathbf{shift}$, to flip input images ($\{1, 0\}$)
\item $\mathbf{flip}$, to shift randomly input images
\end{list}
and real type parameters are:

\vspace{0mm}
\begin{list}{\textbullet}{%
\setlength{\topsep}{1mm}\setlength{\itemsep}{-1mm}}
\item $\mathbf{mu}$, learning rate
\item $\mathbf{mmu}$, momentum rate
\item $\mathbf{l2}$, l2 regularization (weight decay)
\item $\mathbf{l1}$, l1 regularization
\item $\mathbf{maxn}$, maxnorm regularization
\item $\mathbf{drop}$, dropout
\item $\mathbf{noiser}$, noise ratio after activation function
\item $\mathbf{noisesd}$, standard deviation of noise
  ($N(0, \sigma)$)
\item $\mathbf{noiseb}$, ratio of binary noise (only for input layer)
\item $\mathbf{brightness}$, ratio to modify randomly the total brightness of an image
\item $\mathbf{contrast}$, ratio to modify randomly the contrast of an image
\item $\mathbf{lambda}$, to scale the cost factor of an output layer
\end{list}

For data, we can only modify the following (integer) parameter:

\vspace{0mm}
\begin{list}{\textbullet}{%
\setlength{\topsep}{1mm}\setlength{\itemsep}{-1mm}}
\item $\mathbf{balance}$, for balancing data classes
\end{list}

\medskip %
In \ly~we defined a set of functions can be applied to
networks, layers, or data.
\begin{center}
\begin{tabular}{lcl}
$command$&$\rightarrow$&$\mathbf{id}$ ~$.$ ~$\mathbf{id}$ ~$.$ 
          ~$\mathbf{printkernelss}$ ~$($ 
          ~$\mathbf{nfile}$ ~$)$\\[0.5mm]
         &$|$&$\mathbf{train}$ ~$($ ~$\mathbf{cte}$ ~$,$ 
          ~$\mathbf{cte}$ ~$rtrain$ ~$)$\\[0.5mm]
         &$|$&$\mathbf{id}$ ~$.$ ~$\mathbf{train}$ ~$($ 
          ~$\mathbf{cte}$ ~$)$\\[0.5mm]
         &$|$&$\mathbf{id}$ ~$.$ ~$\mathbf{test}$ ~$($ 
          ~$odata$ ~$)$\\[0.5mm]
         &$|$&$\mathbf{id}$ ~$.$ ~$\mathbf{load}$ ~$($ 
          ~$\mathbf{nfile}$ ~$)$\\[0.5mm]
         &$|$&$\mathbf{id}$ ~$.$ ~$\mathbf{save}$ ~$($ 
          ~$\mathbf{nfile}$ ~$)$\\[0.5mm]
         &$|$&$\mathbf{id}$ ~$.$ ~$\mathbf{testout}$ 
          ~$($ ~$\mathbf{nfile}$ ~$)$\\[0.5mm]
         &$|$&$\mathbf{id}$ ~$.$ ~$\mathbf{zscore}$ ~$($ 
          ~$odata$ ~$)$\\[0.5mm]
         &$|$&$\mathbf{id}$ ~$.$ ~$\mathbf{center}$ ~$($ 
          ~$odata$ ~$)$\\[0.5mm]
         &$|$&$\mathbf{id}$ ~$.$ ~$\mathbf{yuv}$ ~$($ 
          ~$)$\\[0.5mm]
         &$|$&$\mathbf{id}$ ~$.$ ~$\mathbf{div}$ ~$($ 
          ~$\mathbf{cte}$ ~$)$\\[0.5mm]
$rtrain$ &$\rightarrow$&$rtrain$ ~$,$ ~$\mathbf{id}$ ~~$|$ 
          ~~$\epsilon$\\[0.5mm]
$odata$  &$\rightarrow$&$\mathbf{id}$ ~~$|$ ~~$\epsilon$\\[0.5mm]
\end{tabular}
\end{center}

For networks the following functions are defined: $\mathbf{train}$,
~to train a network with a specified number of epochs;
$\mathbf{test}$, ~to test a particular network;
$\mathbf{save}$, ~to save the parameters of a network to a particular
file; $\mathbf{load}$, ~to load the parameters of a network from a
particular file; and $\mathbf{testout}$ ~to dump the output of all the
test data to a particular file.

For layers only the next function is defined: $\mathbf{printkernels}$,
~to save the parameters of a layer to a particular file.

For data the following functions are defined: $\mathbf{zscore}$, ~to
normalize data; $\mathbf{yuv}$, ~to convert RGB maps to YUV maps;
$\mathbf{center}$ ~to center data (mean$=0$); and $\mathbf{div}$ ~to
divide all the data by a specific value.

Finally we also define a function that allows us to train together a
list of previously defined networks, specifying the number of epochs
and the number of batches.

\subsection{A compiler for \ly~}
\label{subsec:com}

From the lexical specification, and the syntactic-semantic
specification we have developed a complete
compiler~\cite{AhoUllman2008,Cooper2012} for the specification
language \ly~.
The front-end of this compiler consists of two main modules. The first
one is a scanner to check whether the input program complies with
lexical restrictions defined in Section~\ref{subsubsec:lex}.
This scanner has been implemented using a standard GNU tool for
automatic generation of lexical analyzers:
\texttt{Flex}~\cite{flex08}.

The second module is a parser to check whether the input program
complies with syntactic-semantic constraints defined in
Section~\ref{subsubsec:syn}.
This parser has been implemented using a standard GNU tool for
automatic generation of parsers: \texttt{Bison}~\cite{bison14}.

This front-end is completed with a set of functions representing the
semantic actions necessary to produce intermediate code interpretable
by the \ly~toolkit (the back-end of compiler).

\bibliographystyle{plain}
\bibliography{./parser}

\begin{thebibliography}{1}

\bibitem{AhoUllman2008}
Alfred~V. Aho, Monica~S. Lam, Ravi Sethi, and Jeffrey~D. Ullman.
\newblock {\em Compilers: Principles, Techniques, and Tools (2nd Edition)}.
\newblock Addison Wesley, 2008.

\bibitem{bison14}
GNU Bison.
\newblock A parser generator.
\newblock https://www.gnu.org/software/bison/, 2014.

\bibitem{Cooper2012}
Keith Cooper and Linda Torczon.
\newblock {\em Engineering a Compiler}.
\newblock Morgan Kaufman, 2012.

\bibitem{flex08}
Flex.
\newblock The fast lexical analyzer.
\newblock http://flex.sourceforge.net/, 2008.

\end{thebibliography}
\end{document}